\documentclass[sigconf]{acmart}
\usepackage{multirow}
\usepackage{subfigure}
\usepackage{blindtext}
\usepackage{enumitem}
\usepackage{algorithm}  
\usepackage{algpseudocode}  
\usepackage{amsmath}  
\usepackage{hyperref}
\renewcommand{\algorithmicrequire}{\textbf{Input:}}  
\renewcommand{\algorithmicensure}{\textbf{Output:}} 
\newcommand{\blue}[1]{\textcolor{black}{#1}}
\newcommand{\red}[1]{\textcolor{black}{#1}}
\renewcommand\footnotetextcopyrightpermission[1]{}

\AtBeginDocument{%
  \providecommand\BibTeX{{%
    \normalfont B\kern-0.5em{\scshape i\kern-0.25em b}\kern-0.8em\TeX}}}

\setcopyright{acmcopyright}
\copyrightyear{2022}
\acmYear{2022}
\acmDOI{10.1145/3485447.3512109}

\acmConference[WWW '22] {Proceedings of the ACM Web Conference 2022}{April 25--29, 2022}{Virtual Event, Lyon, France.}
\acmBooktitle{Proceedings of the ACM Web Conference 2022 (WWW '22), April 25--29, 2022, Virtual Event, Lyon, France}
\acmPrice{15.00}
\acmISBN{978-1-4503-9096-5/22/04}

\begin{document}
\fancyhead{}
\title{Cross DQN: Cross Deep Q Network for Ads Allocation in Feed}



\author{Guogang Liao$^{1\dagger}$, Ze Wang$^{1\dagger}$, Xiaoxu Wu$^1$, Xiaowen Shi$^1$, Chuheng Zhang$^{2\ddagger}$, Yongkang Wang$^1$, Xingxing Wang$^1$, Dong Wang$^1$}

\affiliation{%
 \institution{$^1$Meituan, Beijing, China \ \ \ \  $^2$IIIS, Tsinghua University, Beijing, China}
 \city{}
 \country{}
}
\email{{liaoguogang, wangze18, wuxiaoxu04, shixiaowen03}@meituan.com, 
zhangchuheng123@live.com,}
\email{{wangyongkang03, wangxingxing04, wangdong07}@meituan.com}

\renewcommand{\shortauthors}{Guogang Liao and Ze Wang, et al.}

\begin{abstract}
  \renewcommand{\thefootnote}{\fnsymbol{footnote}}
\footnotetext[2]{Guogang Liao and Ze Wang are the corresponding authors.}
\renewcommand{\thefootnote}{\fnsymbol{footnote}}
\footnotetext[3]{This work was done when Chuhang Zhang was an intern in Meituan.}
  \renewcommand{\thefootnote}{\arabic{footnote}}

    E-commerce platforms usually display a mixed list of ads and organic items in feed.
    One key problem is to allocate the limited slots in the feed to maximize the overall revenue as well as improve user experience, which requires a good model for user preference. Instead of modeling the influence of individual items on user {behaviors}, the \textbf{arrangement signal} models the influence of the arrangement of items and may lead to a better allocation strategy. However, most of previous strategies fail to model such a signal and therefore result in suboptimal performance. 
    \blue{In addition, the percentage of ads exposed (PAE) is an important indicator in ads allocation. Excessive PAE hurts user experience while too low PAE reduces platform revenue. Therefore, how to constrain the PAE within a certain range while keeping personalized recommendation under the PAE constraint is a challenge.}

    In this paper, we propose Cross Deep Q Network (Cross DQN) to extract the crucial arrangement signal by crossing the embeddings of different items and \blue{modeling the crossed sequence by multi-channel attention. Besides, we propose an auxiliary loss for batch-level constraint on PAE to tackle the above-mentioned challenge.}
    Our model results in higher revenue and better user experience than state-of-the-art baselines in offline experiments.
    Moreover, our model demonstrates a significant improvement in the online A/B test and has been fully deployed on Meituan feed to serve more than 300 millions of customers.
\end{abstract}


\maketitle

\section{Introduction}
Feed, mixed with organic items and ads, is a popular product on many e-commerce platforms nowadays \cite{Ghose2009AnEA}. Platforms serve users and gain revenue via feed. In general, there are two ways for platforms to get revenue. Firstly, once users consume organic items or ads, the e-commerce platform will gain the platform service fee (hereinafter referred to as fee) according to the orders. Secondly, as an ad is clicked by a user, the platform will charge the corresponding advertiser. For the sake of the platform, displaying more ads is beneficial to ads revenue but harmful to fee since ads are less likely engaging than organic items \cite{Zhao2018ImpressionAF}. Usually, the number of ads is limited in feed to ensure good user experience and engagement. Hence, how to allocate limited slots reasonably and effectively to maximize overall revenue has become a very meaningful and challenging problem \cite{Wang2011LearningTA,Mehta2013OnlineMA,zhang2018whole}.

\begin{figure}[htb]
  \centering
  \includegraphics[width=\linewidth]{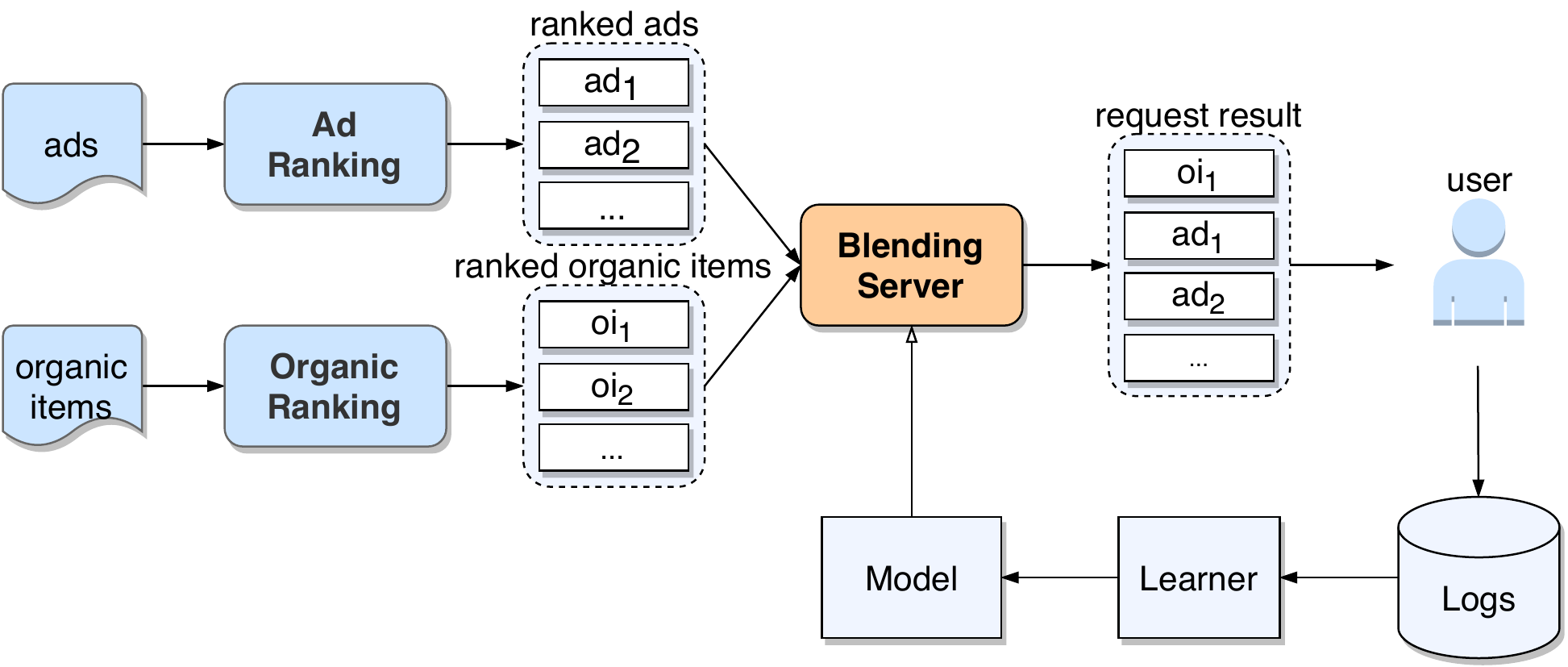}
  \Description{Structure of an ads allocation system. The process of ads allocation takes place in the Blending Server.}
  \caption{Structure of an ads allocation system. The process of ads allocation takes place in the Blending Server.}
  \label{fig:1}
\end{figure}

The structure of an industrial ads allocation system is shown in Figure \ref{fig:1}. Blending Server takes ads sequence and organic items sequence as input and outputs a mixed sequence of the two. For Blending Server, there are two common strategies:
fixed slots strategy and dynamic slots strategy. Most platforms simply allocate ads to pre-determined slots \cite{ouyang2020minet,li2020deep}. Such strategies may lead to suboptimal overall performance. Dynamic slots strategy adjusts the number and slots of ads according to the interest of users. For instance, if a user has a higher tendency to consume commercial ads, the platform will allocate more ads at conspicuous slots to maximize possible benefits. Except for personalization, dynamic slots strategies have lower ads blindness \cite{yan2020LinkedInGEA} and better adaptability, significantly outperforming fixed slots strategy and gradually becoming today's trend. 
Early dynamic slots strategies use some classic algorithms (e.g., Bellman-Ford, unified rank score) to allocate ads slots.
Since the feed is presented to the user in a sequence, recent dynamic ads allocation strategies usually model the problem as Markov Decision Process \cite{sutton1998introduction} and solve it using reinforcement learning (RL) \cite{zhang2018whole,Feng2018LearningTC,zhao2020jointly}.
\begin{figure}[tb]
    \centering
    \includegraphics[width=0.8\linewidth]{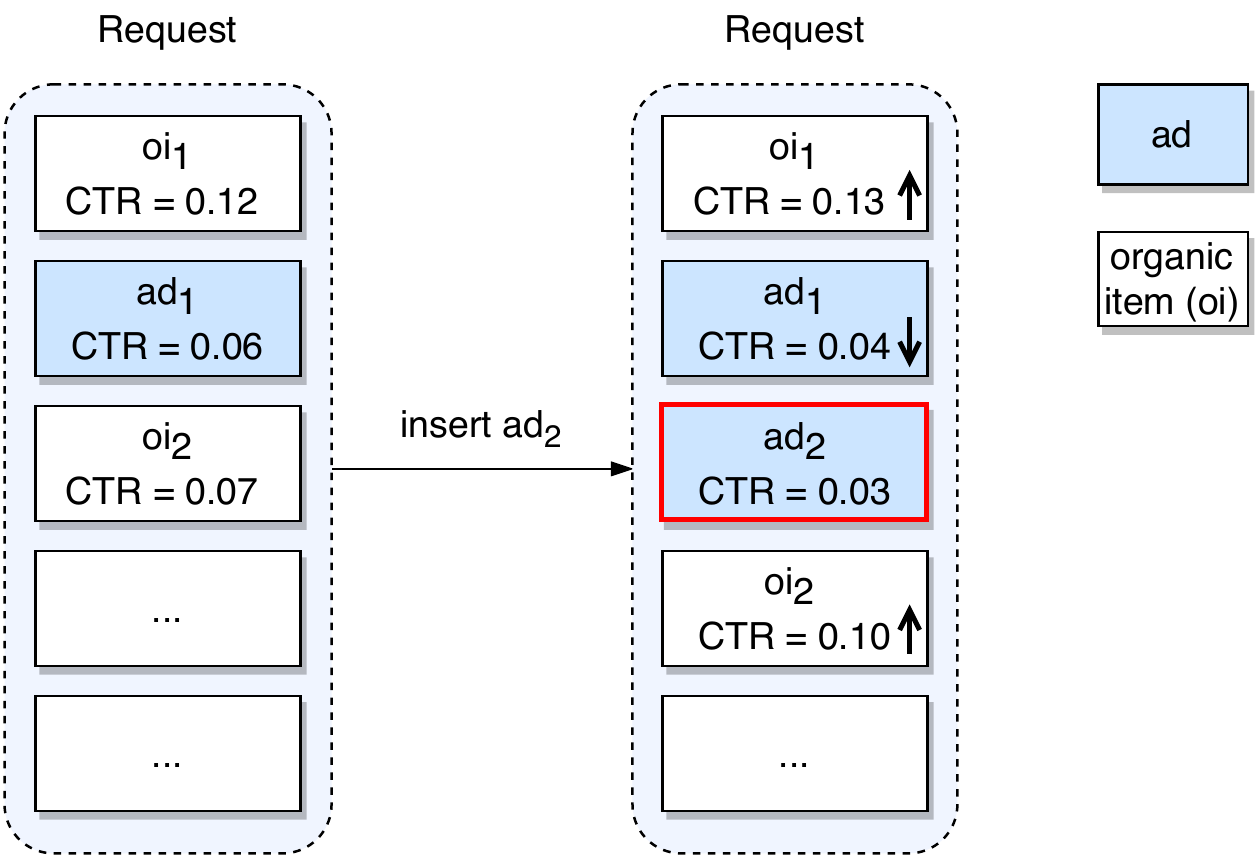}
    \Description{While inserting $ad_{2}$ into feed, the CTR of organic items increases while the CTR of $ad_1$ decreases.}
    \caption{While inserting $ad_{2}$ into feed, the CTR of organic items increases while the CTR of $ad_1$ decreases.}
    \label{fig:2}
  \end{figure}

However, existing RL-based dynamic slots strategies encounter several major limitations: 
i) Most approaches ignore the crucial \textbf{arrangement signal} which is the influence of the arrangement of displayed items on user behaviors. 
For example, as illustrated in Figure \ref{fig:2}, once an ad is inserted into feed, the click-through rate (CTR) of surrounding organic items and ads fluctuate.  This signal receives attention in the scenario of Re-Rank recently \cite{Carrion2021BlendingAW, Wei2020GeneratorAC, Feng2021RevisitRS,Feng2021GRNGR} but is largely neglected in ads allocation.
ii) Most of existing methods lack an efficient balance between the personalization of different requests and the constraint on the percentage of ads exposed (PAE) in a period. 
PAE is the most important constraint in ads allocation, which balances the user experience and platform revenue. 
Previous methods constrain all the requests or the requests within the same hour \cite{wang2019DDPGCHER} with the same target PAE, resulting in a lack of personalization and differentiation in the allocation of ads between different requests.

To address the limitations of existing methods, we present a novel framework called \emph{Cross Deep Q Network} (Cross DQN) based on deep reinforcement learning. 
\blue{Specifically, we design two novel units called \emph{State and Action Crossing Unit} (SACU) and \emph{Multi-Channel Attention Unit} (MCAU) to explicitly extract the arrangement signal.}
Besides, we propose an auxiliary loss for batch-level constraint to balance the personalization of different requests and the constraint on PAE in a period.

The contributions of our work are summarized as follows:
\begin{itemize}[leftmargin=*]

  \item \textbf{A superior ads allocation strategy}. 
  In this paper, we propose a novel RL-based framework named Cross DQN\footnote{The code and data example are publicly accessible at \url{https://github.com/weberrr/CrossDQN}.} to dynamically adjust the number and the slots of ads in feed, which can effectively extract the arrangement signal and reasonably balance personalization of different requests and the constraint on PAE.
  
  \item \textbf{Detailed industrial and practical experience}. We successfully deploy Cross DQN on the Meituan feed and obtain significant improvements in both platform revenue and user experience. 
\end{itemize}

  

\section{Related Works}

Traditional strategy for ads allocation in feed is to display ads at fixed slots. Recently, dynamic ads allocation strategies gains growing attention. According to whether RL is used, existing dynamic ads allocation strategies can be roughly categorized into two {categories}: non RL-based and RL-based.

Non RL-based methods {usually} use classical algorithms to allocate ads slots. \citet{koutsopoulos2016optimal} define ads allocation as a shortest-path problem on a weighted directed acyclic graph {where nodes represent ads or slots and edges represent expected revenue.}
The shortest path can be found by running Bellman-Ford algorithm.
Furthermore, \citet{yan2020LinkedInGEA} takes the impact of the interval between ads into consideration and re-ranks ads and organic items jointly via a uniform {ranking} formula.

RL-based methods model the ads allocation problem as an MDP and solved it with different RL techniques. \citet{zhao2020jointly} proposes a two-level RL framework to jointly optimize the recommending and advertising strategies. \citet{zhao2019deep} proposes a DQN architecture to determine the optimal ads and ads position jointly. 
\citet{xie2021hierarchical} proposes a hierarchical RL-based framework to
first decide the channel and then determine the specific item for each slot.
In contrast to the previous work, we incorporate the arrangement signal into a RL-based dynamic ads allocation model to improve the performance.

\section{Problem Formulation}
\label{sec:problem}
In our scenario, we present $K$ slots in one screen and handle the allocation for each screen in the feed of a request sequentially.
The ads allocation problem is formulated as a Constrained Markov Decision Process (CMDP) \cite{altman1999constrained} ($\mathcal{S}$, $\mathcal{A}$, $r$, ${P}$, $\gamma$, $\mathcal{C}$), the elements of which are defined as follows:

\begin{itemize}
    \item \textbf{State space $\mathcal{S}$}. A state $s \in \mathcal{S}$ {consists of the information of candidate items (i.e., the ads sequence and the organic items sequence which are available on current step $t$), the user (e.g., age, gender and historical behaviors), and the context (e.g., order time).} 
    \item \textbf{Action space $\mathcal{A}$}. An action $a \in \mathcal{A}$ is the decision whether to display an ad on each slot on the current screen, which is formulated as follows:
    \begin{equation}
        a=(x_{1}, x_{2}, \ldots, x_{K}),
    \end{equation}
    where $x_{k}=\begin{cases}
        1 &\text{display an ad in the $k$-th slot} \\ 
        0 &\text{otherwise} 
        \end{cases}$, $\forall k \in [K]$.
    In our scenario, {we do not change the order of ads sequence and organic items sequence in Blending Server}.
    \item \textbf{Reward $r$}. After the system {takes} an action in one state, a user browses the mixed list and gives a feedback. {The reward is calculated based on the feedback and consists of} 
    ads revenue $r^\text{ad}$, fee $r^\text{fee}$ and user experience $r^\text{ex}$:
    \begin{equation}
      \begin{aligned}
    r(s,a) &= r^\text{ad}+r^\text{fee}+\eta r^\text{ex}, \\
      \text{where}\ \ r^\text{ex}& = \begin{cases}
        2& \text{click and order}\\
        1& \text{click and leave}\\
        0& \text{no click and leave}
      \end{cases}.\end{aligned}
      \label{eq:eta}
    \end{equation}
    \item \textbf{Transition probability ${P}$}.
    $P(s_{t+1}|s_t,a_t)$ is defined as the state transition probability from $s_t$ to $s_{t+1}$ after taking the action $a_t$, where $t$ is the index for the screen/time step.
    The action taken in the state affects user behaviors. 
    When the user pulls down, the state $s_{t}$ transits to the state of next screen $s_{t+1}$. 
    The items selected to present by $a_t$ will be removed from the state on the next step $s_{t+1}$.
    If the user no longer pulls down, the transition terminates. 
    
    \item \textbf{Discount factor $\gamma$}. The discount factor $\gamma \in [0, 1]$ balances the short-term and long-term rewards.
    \item \textbf{Constraint $\mathcal{C}$}. 
    The platform-level constraint is that the absolute difference between the total PAE in a period and the the target value $\delta$ {should be less} than a threshold $\varepsilon$ {to ensure stable ads revenue}.
    The PAE is formulated as follows:
    \begin{equation}
      \begin{aligned}
       \text{PAE} = \frac{\sum_{1 \leq i \leq N}\text{Num}^\text{ad}_{i}}{\sum_{1 \leq i \leq N}(\text{Num}^\text{ad}_{i} + \text{Num}^\text{oi}_{i})}\ ,
      \end{aligned}
    \end{equation}
    where $N$ is the number of requests in a period, $\text{Num}^\text{ad}_{i}$ and $\text{Num}^\text{oi}_{i}$ mean the number of ads and organic items in the $i$-th request. In this work, we choose one week as {the} period. 
    Consequently, the platform-level constraint can be formulated as follows:
    \begin{equation}
      |\text{PAE} - \delta| < \varepsilon.
    \end{equation}
    
  \end{itemize}
   
  {Given the CMDP formulated as above, the objective is to find an ads allocation policy $\pi: \mathcal{S} \rightarrow \mathcal{A}$ to maximize the total reward under the platform-level constraint.}

  \section{Methodology}
\begin{figure}[b]
  \centering
  \subfigure[Dueling DQN \ \ ]{
  \begin{minipage}[t]{0.5\linewidth}
  \centering
  \includegraphics[height=1.46in]{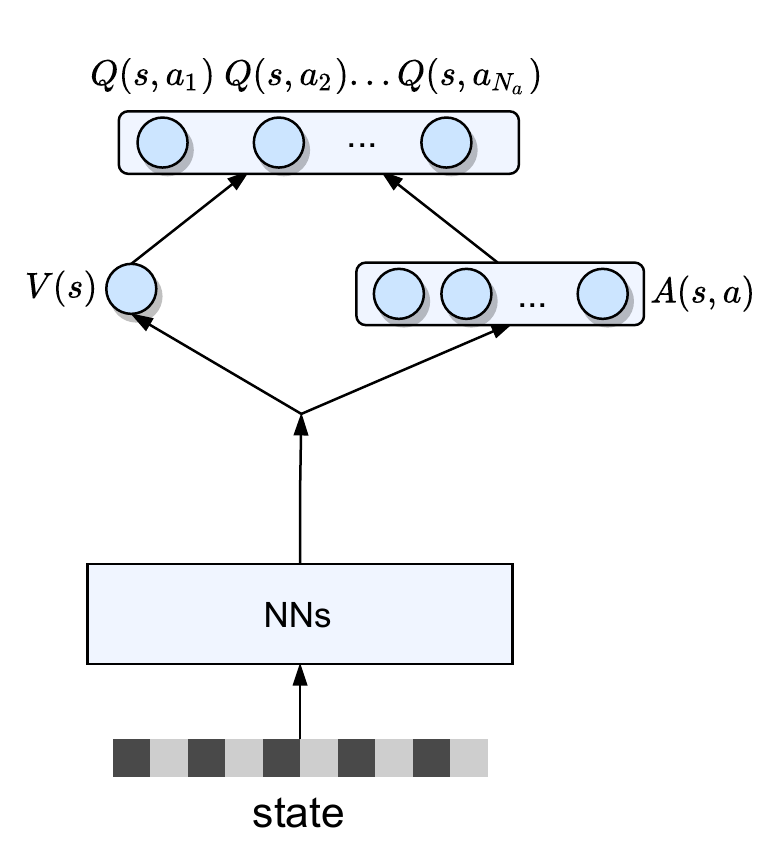}
  \label{fig:subfig:duelingDQN} 
  \end{minipage}%
  }%
  \subfigure[Cross DQN]{
  \begin{minipage}[t]{0.5\linewidth}
  \centering
  \includegraphics[height=1.46in]{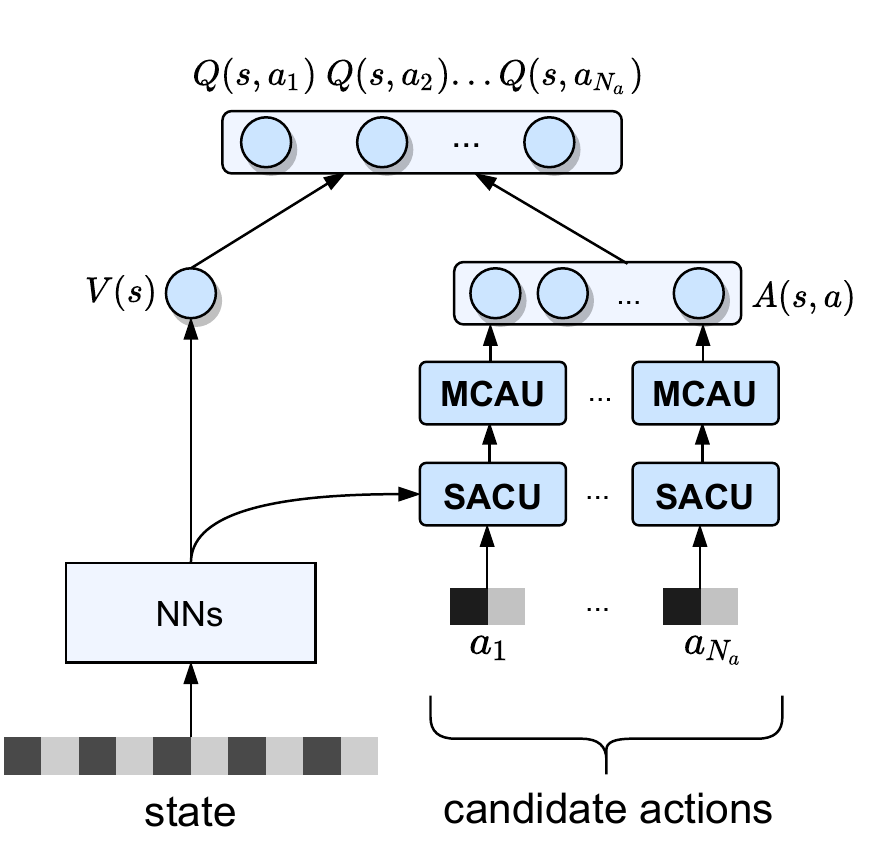}
  \label{fig:subfig:mcDQN} 
  \end{minipage}%
  }%
  \caption{Architectures of Dueling DQN and Cross DQN.}
  \Description{Architectures of Dueling DQN and Cross DQN.}
  \label{fig:2DQN}
\end{figure}

    \begin{figure*}[tb]
      \centering
      \includegraphics[width=\textwidth]{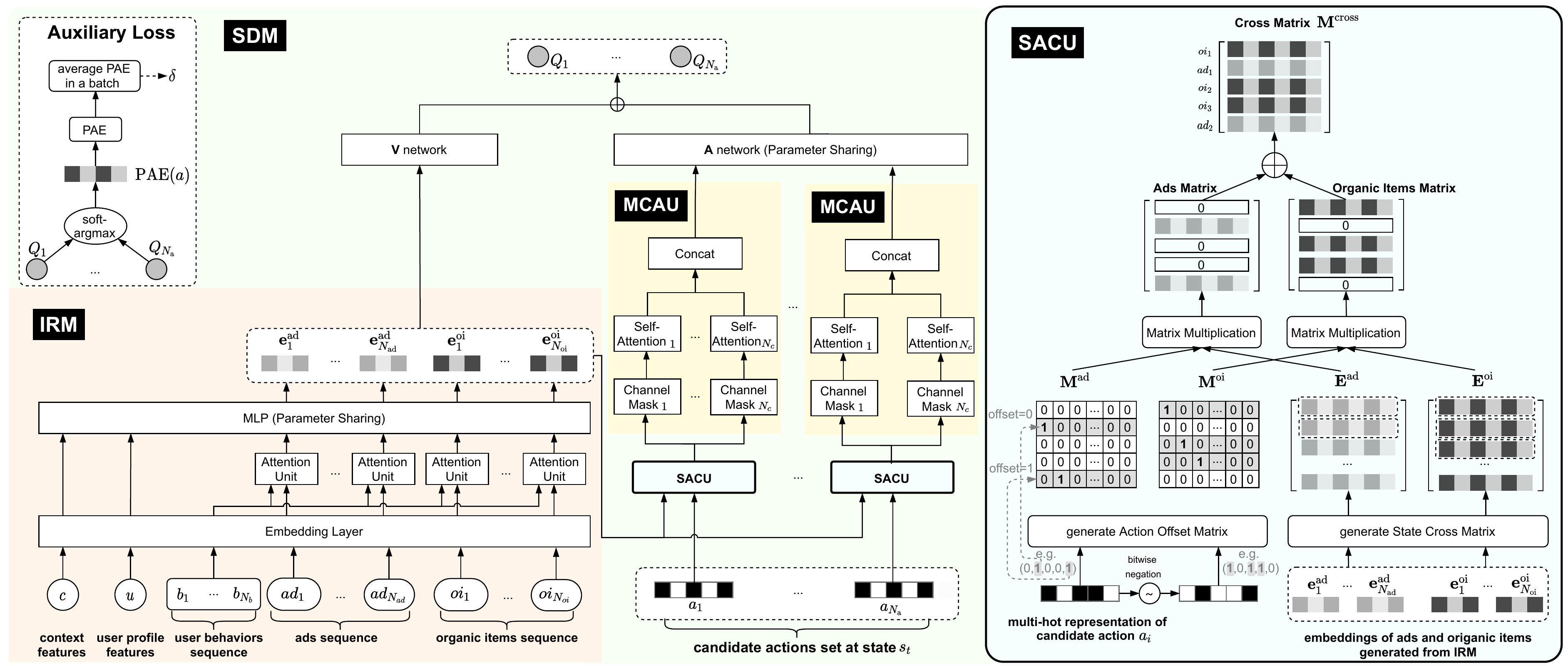}
      \Description{Network architecture of Cross DQN.
      Item Representation Module (IRM) generates the state embedding based on the raw state.
      Sequential Decision Module (SDM) generates Q-values of different actions with the help of State and Action Crossing Unit (SACU), Multi-Channel Attention Unit (MCAU) and auxiliary loss for batch-level constraint.}
      \caption{
      Network architecture of Cross DQN.
      Item Representation Module (IRM) generates the state embedding based on the raw state.
      Sequential Decision Module (SDM) generates Q-values of different actions with the help of State and Action Crossing Unit (SACU), Multi-Channel Attention Unit (MCAU) and auxiliary loss for batch-level constraint.
      }
      \label{fig:mcDQN}
    \end{figure*}
  
  \subsection{Architecture Overview}
  The structure of {popular RL model,} Dueling DQN, \cite{wang2016duelingDQN} is {shown} in Figure \ref{fig:subfig:duelingDQN}, {which receives the state only as the input.}
  Such a structure fails to extract cross information between the action and the state, making it difficult to model the arrangement signal of mixed lists. 
  A common solution is to concatenate state and action directly, 
  {but the model remains hard to extract the information between the items in the mixed list designated by the action.}
  {To this end,}
  we propose a novel structure Cross DQN (cf. Figure \ref{fig:subfig:mcDQN}). 
  The State and Action Crossing Unit (SACU) is designed to cross the embeddings of
  the items in a mixed list designated by the action. \blue{The Multi-Channel Attention Unit (MCAU) is designed to effectively extract arrangement signal from different channel combinations.}

Specifically, we show the detailed structure of Cross DQN in Figure \ref{fig:mcDQN}.
The model takes a state (including the organic items/ads sequence, context information, etc.) and the corresponding candidate actions as the input.
Then, Item Representation Module (IRM) generates the
representations (especially the representations of ads and organic items).
Next, Sequential Decision Module (SDM) generates Q-values of different actions with the help of SACU, \blue{MCAU} and an auxiliary loss for batch-level constraint.
In SACU, the state embeddings are intersected according to the action to form a unified matrix representation.
\blue{In MCAU, the crossing matrix generated from SACU are split into different channels to calculate the multi-channel attention weight.}
Finally, SDM chooses the action with the largest Q-value.
We will introduce them in detail in the following subsections.

\subsection{Item Representation Module}

Item Representation Module (IRM) generates the state embedding from the raw state. 
To efficiently process the information from different sources, IRM generates two sequences of mixed embeddings: one for ads and one for organic items.
The embedding for each item encodes not only the information of the item itself but also the information of the user profile, the context, and the interaction with historical user behaviors.

First, we use embedding layers to extract the embeddings from raw inputs.
We denote the embeddings for ads, organic items, historical behaviors of the user, the user profile, the context as $\{\mathbf{e}_i^\text{ad}\}_{i=1}^{N_\text{ad}}$, $\{\mathbf{e}_i^\text{oi}\}_{i=1}^{N_\text{oi}}$, $\{\mathbf{e}_i^\text{b}\}_{i=1}^{N_\text{b}}$, $\mathbf{e}^\text{u}$, and $\mathbf{e}^\text{c}$ respectively, where the subscript $i$ denotes the index within the sequence and $N_\text{ad}$, $N_\text{oi}$, and $N_\text{b}$ are the number of ads, organic items, and historical behaviors.
Then, 
we use a target attention unit \cite{vaswani2017attention} to encodes the interaction between the historical behaviors of the user and the corresponding item, which is similar to \citet{zhou2018DIN}:
\begin{equation}
  \begin{aligned}
  \mathbf{e}_j^\text{ad}& \leftarrow \text{Att} \Big( \mathbf{e}^\text{ad}_{j}, \{\mathbf{e}^\text{b}_{i} \}_{i=1}^{N_\text{b}} \Big), \forall j\in [N_\text{ad}]; \\
  \mathbf{e}_j^\text{oi}& \leftarrow \text{Att} \Big( \mathbf{e}^\text{oi}_{j}, \{\mathbf{e}^\text{b}_{i} \}_{i=1}^{N_\text{b}} \Big), \forall j\in [N_\text{oi}].
\end{aligned}
\end{equation}

Afterwards, we append the embeddings of the user profile and the context to the embedding of each item: 
  
\begin{equation}
  \begin{aligned}
  \mathbf{e}_j^\text{ad}& \leftarrow \text{MLP}\Big( \mathbf{e}^\text{ad}_{j} || \mathbf{e}^\text{u}|| \mathbf{e}^\text{c} \Big), \forall j\in [N_\text{ad}];  \\
    \mathbf{e}_j^\text{oi}& \leftarrow \text{MLP}\Big( \mathbf{e}^\text{oi}_{j} || \mathbf{e}^\text{u}|| \mathbf{e}^\text{c} \Big), \forall j\in [N_\text{oi}];
  \end{aligned}
\end{equation}
where $||$ denotes concatenation.
Notice that there are some strong features for ads and organic items in our scenario (e.g., discount, delivery fee, delivery time), which are concatenated with the embedding of each item and input into SDM.

For ease of notation, we can also write the embeddings for ads and organic items in matrix form, each row of which represents one item in the sequence, i.e., 
\begin{equation}
\begin{aligned}
    \mathbf{E}^\text{ad} & = \Big[ \mathbf{e}^\text{ad}_{1}\Big|\mathbf{e}^\text{ad}_{2}\Big|...\Big|\mathbf{e}^\text{ad}_{N_\text{ad}}\Big]^T, \\
     \mathbf{E}^\text{oi} & = \Big[ \mathbf{e}^\text{oi\;\;\;\!\!\!\!}_{1}\Big|\mathbf{e}^\text{oi\;\;\;\!\!\!\!}_{2}\Big|...\Big|\mathbf{e}^\text{oi\;\;\;\!\!\!\!}_{N_\text{oi\;\;\;\!\!\!\!\;\;\!\!\!}}\Big]^T.
\end{aligned}
\end{equation}

Involving several attention units, IRM may be time-consuming upon deployment. 
However, IRM is an independent module within Cross DQN so that we can invoke IRM in parallel to other modules preceding Cross DQN. See more details in Section \ref{sec:online imp}.

\subsection{State and Action Crossing Unit}
\label{sec:SACU}

To evaluate the Q-value of a certain state-action pair, we need an efficient representation of the mixed list designated by the corresponding action.
State and Action Crossing Unit (SACU) helps us to construct a sequence of embeddings corresponding to the mixed list from the state embedding.

First, 
given an action, 
we generate the corresponding action offset matrices for ads and organic items:
$\mathbf{M}^\text{ad} \in \{0, 1\}^{K\times N_\text{ad}}$ 
and $\mathbf{M}^\text{oi} \in \{0, 1\}^{K\times N_\text{oi}}$, 
where the $(i,j)$-th element represents whether the $j$-th ad/organic item is presented on the $i$-th slot.
Recall that $K$ is the number of slots in one screen.
For example, given the action $a=(0,1,0,0,1)$ with $K=5$, the action offset matrix $\mathbf{M}^\text{ad}$ is 
\begin{equation}
    \mathbf{M}^{\text{ad}} = 
    \left[
    \begin{array}{ccccc}
        0 & 0& \dots & 0 & 0 \\
        1 & 0& \dots & 0 & 0 \\
        0 & 0& \dots & 0 & 0 \\
        0 & 0& \dots & 0 & 0 \\
        0 & 1& \dots & 0 & 0
    \end{array}
    \right] 
    \in \mathbb{R}^{K \times N_{\text{ad}}}.       
    \end{equation} 

Then, we can calculate the cross matrix $\mathbf{M}^\text{cross}$, which is the embedding of the mixed list corresponding to the given action, using the action offset matrices.
  \begin{equation}
    \mathbf{M}^\text{cross}= \mathbf{M}^\text{ad} \mathbf{E}^\text{ad} +\mathbf{M}^\text{oi} \mathbf{E}^\text{oi}.
  \end{equation}

With SACU, we generate the embedding for the mixed list which enables us to efficiently extract the arrangement signal \blue{in the next module}.

\subsection{\blue{Multi-Channel Attention Unit}}
\blue{The user may focus on one or more aspects (e.g., discount, delivery fee, delivery time) of the mixed sequence at the same time. Accordingly, we propose MCAU to simultaneously model the user's attention to different aspects of the mixed sequence.}

\blue{The cross matrix $\mathbf{M}^\text{cross} \in \mathbb{R}^{K \times N_{\text{e}}}$ generated by SACU contains $N_{\text{e}}$ different channels. Each channel represents an information dimension in the latent space and can be used to model one aspect of the mixed sequence. Meanwhile, the user may pay attention to more than one aspect of the mixed sequence at the same time. So the sequence information of two or more channels need to be combined for modeling. Next we will detail that how the sequence information of multiple channels is combined and modeled.}

\blue{For $N_\text{e}$ channels, we formulate the number of channel combinations as $N_\text{c}$, which is calculated as follows:}
\begin{equation}
  {N_\text{c} = 2^{N_\text{e}}-1.}
\end{equation}

\blue{We formalize the mask matrix for the $i$-th combination as $\mathbf{M}^\text{mask}_i$. For example, the mask matrix for combination of the first channel and the last channel is} 
\begin{equation}
  \blue{
  \mathbf{M}^{\text{mask}}_i = 
  \left[
  \begin{array}{ccccc}
      1 & 0& \dots & 0 & 1 \\
      1 & 0& \dots & 0 & 1 \\
      1 & 0& \dots & 0 & 1 \\
      1 & 0& \dots & 0 & 1 \\
      1 & 0& \dots & 0 & 1
  \end{array}
  \right] 
  \in \mathbb{R}^{K \times N_{\text{e}}}.       
  }
  \end{equation}

  \blue{Nextly, signal matrix calculated by cross matrix and mask matrix are input into corresponding self-attention network \cite{vaswani2017attention} to model the attention across the $K$ items and generate a latent vector, as follows:}
\begin{equation}
  \blue{\mathbf{M}^\text{signal}_{i} = \mathbf{M}^\text{cross} \odot \mathbf{M}^\text{mask}_i,\ \forall i \in [{N_\text{c}}].}
\end{equation}
  \begin{equation}
    \blue{\mathbf{e}^\text{signal}_{i} = \text{flatten}\Big( \text{SelfAtt}_{(i)}\big(\mathbf{M}^\text{signal}_{i} \big) \Big),\ \forall i \in [{N_\text{c}}].}
\end{equation}

\blue{Latent vectors output by different self-attention network are concatenated together to represent the arrangement signal extracted from different channels, as follows:}
\begin{equation}
  \blue{\mathbf{e}^\text{signal} = \mathbf{e}^\text{signal}_{1}|| \mathbf{e}^\text{signal}_{2}||...||\mathbf{e}^\text{signal}_{{N_\text{c}}}.}
\end{equation}

\subsection{Sequential Decision Module}
\label{sec:SDM}

With the help of SACU \blue{and MCAU}, Sequential Decision Module (SDM) takes the embeddings generated by IRM and candidate actions as the input, and outputs Q-values corresponding to different actions.
  
Given a set of $N_\text{a}$ candidate actions $\{a_i\}_{i=1}^{N_\text{a}}$, SACU generates a cross matrix $\mathbf{M}^\text{cross}_i$ for each action \blue{and MCAU generates corresponding arrangement signal representation for each action.} 
  Subsequently, the outputs of the V network and the A network \cite{wang2016duelingDQN} can be calculated as follows:
  \begin{equation}
    V(s) = \text{MLP} \big(\text{flatten}(\text{pool}(\mathbf{E}^\text{ad}) || \text{pool}(\mathbf{E}^\text{oi}) )) \big).
  \end{equation}
  \begin{equation}
    A(s,a_{i}) =\text{MLP}\big(\mathbf{e}^\text{\blue{signal}}_{i}\big),
  \end{equation}
where pool indicates average pooling over different rows (i.e., different items in the ads/organic items sequence).
  
Finally, SDM outputs the Q-value $Q(s,a_{i})$ corresponding to the $i$-th candidate action on the current screen as follows:
\begin{equation}
    \label{eq:dulDQN}
    Q(s,a_{i}) =V(s)+\Big(A(s, a_{i})-\frac{1}{N_\text{a}}\sum_{j=1}^{N_\text{a}} A(s,a_j)\Big).
\end{equation}

\subsection{Auxiliary Loss for Batch-level Constraint}
\label{sec:aux loss} 

Recall that our objective is to maximize cumulative reward under the constraint on the average ads exposure.
The key for a successful strategy is to satisfy the constraint while maintaining a differentiated recommendation for different users/scenarios.
For example, if the user is prone to be annoyed by ads, we should expose less ads to the user, and vice versa. 
Different auxiliary losses to constrain the percentage of ads exposure (PAE) can result in different level of differentiation.
A common solution is to use a request-level constraint, i.e., constraining the PAE of each request to be close to the PAE target $\delta$.
Such a solution may result in poor differentiation since the PAE of each request is constrained to the same target $\delta$ regardless of the context.
To allow for differentiation, \citet{wang2019DDPGCHER} propose to use an hour-level constraint that allows for using different PAE targets in different hours.
However, the level of differentiation is still limited within an hour.
To this end, we propose a batch-level constraint to constrain the average PAE of the requests in a batch instead of constraining the PAE of each request.

We denote the PAE associated with the action $a$ as $\text{PAE}(a)$. 
For example, the PAE of $a=(0,1,0,0,1)$ is $0.4$.
Given a batch of transitions $B$, our batch-level constraint can be written as:
\begin{equation}
  \label{eq:loss_res}
    L_{{\text{PAE}}}(B)  = \Big( \delta - \frac{1}{|B|} \sum_{s\sim B}  \text{PAE}(\text{arg}\max_{a\in\mathcal{A}} Q(s,a)) \Big)^2.
\end{equation}
However, the argmax function is not differentiable.
Therefore, we use a soft version of argmax instead, i.e., we use 
\begin{equation}
  \label{eq:func_res}
  \text{PAE}(\text{arg}\max_{a\in\mathcal{A}} Q(s,a))  \approx
  \sum_{i=1}^{N_\text{a}} \frac{1}{Z} \exp \big[\beta Q(s,a_i) \big] \text{PAE}(a_i),
\end{equation}
where $Z=\sum_{j=1}^{N_\text{a}} \exp [\beta Q(s,a_j) ]$ is the normalization factor and $\beta$ is the temperature coefficient.

Unlike previous request-level or hour-level constraints that limit the PAE for each request, we only limit the \emph{average} PAE estimated using randomly sampled batches.
Such a weaker form of constraint encourages the model to choose the action with a PAE that is deviated from $\delta$ but may better adapt for the current context.

\subsection{Offline Training}

We show the process of offline training in Algorithm \ref{alg:offline}.
We train Cross DQN based on an offline dataset $D$ generated by an online exploratory policy $\pi_b$.
For each iteration, we sample a batch of transitions $B$ from the offline dataset and update the model using gradient back-propagation w.r.t. the loss:
\begin{equation}
  \label{eq:loss}
  L(B) = L_\text{DQN}(B) + \alpha L_\text{PAE}(B),
\end{equation}
where $L_\text{DQN}$ is the same loss function as the loss in DQN \cite{mnih2015human}, $L_\text{PAE}$ is the auxiliary loss for the constraint, and $\alpha$ is the coefficient to balance the two losses.
Specifically, 
\begin{equation}
  L_\text{DQN} (B) = \frac{1}{|B|} \!  \sum_{(s,a,r,s')\in B} \! \! \Big( r + \gamma \max_{a'\in\mathcal{A}} Q(s', a') - Q(s,a) \Big) ^ 2.
\end{equation} 

\begin{algorithm}[htbp]  
  \renewcommand\arraystretch{1.1}
  \caption{Offline Training of Cross DQN}
  \label{alg:offline}
  \begin{algorithmic}[1] 
    \State Offline data $D=\{(s,a,r,s')\}$ (generated by an online exploratory policy $\pi_b$)
    \State Initialize a value function $Q$ with random weights
    \Repeat
        \State Sample a batch $B$ of $(s,a,r,s')$ from $D$
        \State Update network parameters by minimizing $L(B)$ in \eqref{eq:loss}
    \Until Convergence
  \end{algorithmic}  
\end{algorithm}  

\subsection{\blue{Online Serving}} 
\label{sec:online imp}

We show the process of online serving in Algorithm \ref{alg:online1}. 
In the online serving system, Cross DQN selects
the action with the highest reward 
based on current state and converts the action to ads slots set for the output.
When the user pulls down, the model receives the state for the next screen, and then makes a decision based on the information on the next screen.

\begin{algorithm}[hbtp]  
  \renewcommand\arraystretch{1.1}
\renewcommand{\algorithmicrequire}{\textbf{Input:}}
\renewcommand{\algorithmicensure}{\textbf{Output:}}
    \caption{Online Inference of Cross DQN}
    \label{alg:online1}
    \begin{algorithmic}[1] 
      
      \State Initial state $s_0$
      \Repeat
      \State Generate $a^*_{t} = \arg \max_{a\in \mathcal{A}} Q(s_t,a)$
      \State Allocate ads slots following  $a^*_{t}$
      \State User pulls down to the next screen $t+1$
      \State Observe the next state $s_{t+1}$
       \Until User leaves
    \end{algorithmic}  
  \end{algorithm}  
  
  In our scenario, current state will transit to the next state or terminate depending on whether the user pulls down.
However, the next possible state corresponding to a given action is deterministic if the interaction does not terminate.
Based on this observation, we can cache the decisions for multiple screens and transmit to the client at once to reduce the time cost for the communication between the server and the client.

\begin{figure}[tb]
  \centering
  \includegraphics[width=0.98\linewidth]{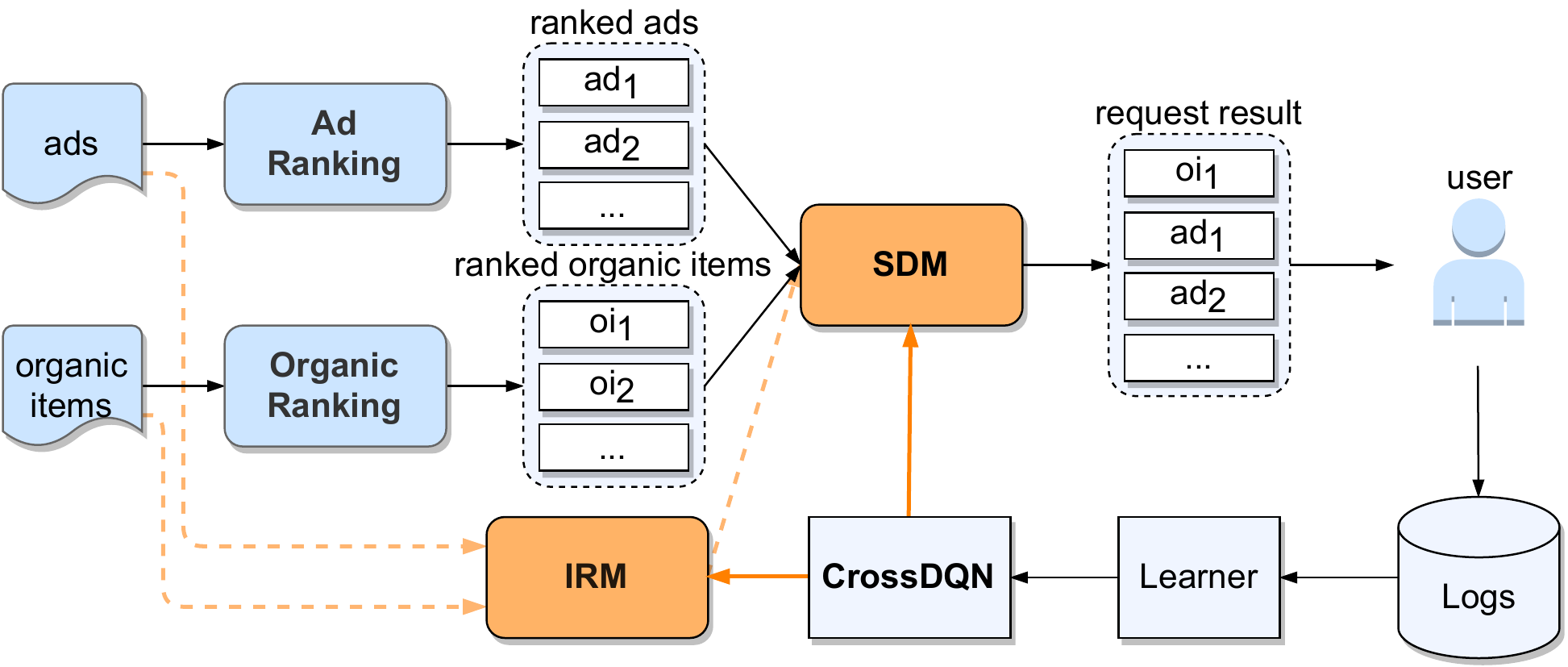}
  \caption{Model decomposition for online service.}
  \Description{Model decomposition for online service.}
  \label{fig:6}
\end{figure}

\textbf{Model Decomposition}. 
Cross DQN will be called $T$ times for one cache, which is time consuming for industrial scenarios where latency is a major concern.
Fortunately, the outputs of IRM can be reused across different calls of Cross DQN in one cache, which saves up to about $80\%$ computation time.
Since the generation of item representations does not rely on the previous modules (such as ranking and ads bidding), we calculate the representation of the items parallel to the previous modules, which further reduces the latency.
As shown in Figure \ref{fig:6}, the Cross DQN is decomposed into IRM and SDM for deployment. The two parts are trained end-to-end but deployed on different services for real-time prediction. The IRM is calculated parallel to Ad Ranking and Organic Ranking systems. Hence, it is latency-free for real-time inference of SDM.

\label{sec:param sharing} 
\textbf{Parameter Sharing}. Both of IRM and SDM use parameter sharing (cf. Figure \ref{fig:mcDQN}) across different ads/organic items. 
As for IRM, we can calculate the representations for all recalled items at the same time without ranking information
through parameter sharing and parallel computing.
Meanwhile, in SDM, we use parameter sharing across different actions to guarantee the consistency of the reward evaluation of the actions and ensure that the batch-level constraint is effective. 
In addition, parameter sharing can reduce the scale of parameters, accelerate model training and reduce memory usage.

  \section{Experiments}
  We will evaluate
  Cross DQN model through offline and online experiments in this section. 
  In offline experiments, we will compare Cross DQN with existing baselines and analyze the role of different designs in Cross DQN.
  In online experiments, we will compare Cross DQN with the previous strategy deployed on the Meituan platform using an online A/B test.
  
  \subsection{Experimental Settings}
  \begin{table}[bp]
    \Description{Statistics of the dataset.}
    \caption{Statistics of the dataset.}
    \renewcommand\arraystretch{1.3}
    \centering
    \begin{tabular}{ccccc}
      \hline
    \#requests    & \#users  & \#ads   & \#items \\
    \hline
    12,729,509 & 2,000,420 & 385,383 & 726,587 \\
    \hline
    \end{tabular}
    \label{dataset}
  \end{table}
  \subsubsection{Dataset}
We collect the dataset by running an exploratory policy on the Meituan platform during March 2021.
We present the detailed statistics of the dataset in Table \ref{dataset}. 
Notice that each request contains several transitions.
The features for the ads/organic items include the identity, the category, the comment score, etc. 
The features for the user profile include the identity, the gender, etc.

\begin{table*}[tb]

  \Description{The result of Revenue Indicators and Experience Indicators. Each experiment are presented in the form of mean $\pm$ standard deviation. The improvement means the improvements of Cross DQN across the best baselines.}
  \caption{The result of Revenue Indicators and Experience Indicators. Each experiment are presented in the form of mean $\pm$ standard deviation. The improvement means the improvements of Cross DQN across the best baselines.}
  \label{tab:all_res}
  \renewcommand\arraystretch{1.15}
  \setlength{\tabcolsep}{5.7mm}{
  \begin{tabular*}{\textwidth}{c|cc|cc}
  \hline

  \multirow{2}{*}{model}  & \multicolumn{2}{c|}{Revenue Indicators}                       & \multicolumn{2}{c}{Experience Indicators}         \\
                     & $R^\text{ad}$               & $R^\text{fee}$              & $R^\text{cxr}$              & $R^\text{ex}$               \\
  \hline \hline
  Fixed              & 0.2211\ ($\pm$0.00252) & 0.2476\ ($\pm$0.00686) & 0.2148\ ($\pm$0.00342) & 0.8823\ ($\pm$0.00730) \\
  GEA                & 0.2372\ ($\pm$0.00035) & 0.2564\ ($\pm$0.00096) & 0.2457\ ($\pm$0.00061) & 0.9493(\ $\pm$0.00012) \\
  CTLRL                & 0.2286\ ($\pm$0.00101) & 0.2536\ ($\pm$0.00860) & 0.2250\ ($\pm$0.00213) & 0.9078\ ($\pm$0.00384) \\
  HRL-Rec                & 0.2380\ ($\pm$0.00287) & 0.2660\ ($\pm$0.00132) & 0.2530\ ($\pm$0.00021) & 0.9526\ ($\pm$0.00123) \\
  DEAR               & 0.2391\ ($\pm$0.00244) & 0.2687\ ($\pm$0.00116) & 0.2530\ ($\pm$0.00044) & 0.9552\ ($\pm$0.00407) \\
  \hline
  \textbf{Cross DQN}       & \textbf{0.2465\ ($\pm$0.00058)} & \textbf{0.2742\ ($\pm$0.00135)} & \textbf{0.2551\ ($\pm$0.00081)} & \textbf{0.9703\ ($\pm$0.00085)}   \\
  \quad -aux       & 0.2446\ ($\pm$0.00079) & 0.2737\ ($\pm$0.00231) & 0.2537\ ($\pm$0.00034) & 0.9671\ ($\pm$0.00118) \\
  \quad -aux-mcau     & 0.2418\ ($\pm$0.00120) & 0.2728\ ($\pm$0.00102) & 0.2534\ ($\pm$0.00045) & 0.9661\ ($\pm$0.00092) \\
  \quad -aux-mcau-sacu    & 0.2370\ ($\pm$0.00217) & 0.2722\ ($\pm$0.00286) & 0.2508\ ($\pm$0.00065) & 0.9629\ ($\pm$0.00201) \\
  \hline \hline
  Improvement        & \textbf{3.09\%}      & \textbf{2.05\%}      & \textbf{0.83\%}      & \textbf{1.58\%}      \\
  \hline
  \end{tabular*}}
  \label{tab:all_res}
\end{table*}

\begin{figure*}[tb] 
    \centering 
    \subfigure[{The reward curve of $\eta$.}]{ 
      \centering
      \begin{minipage}{0.337\linewidth}
      \includegraphics[width=\textwidth]{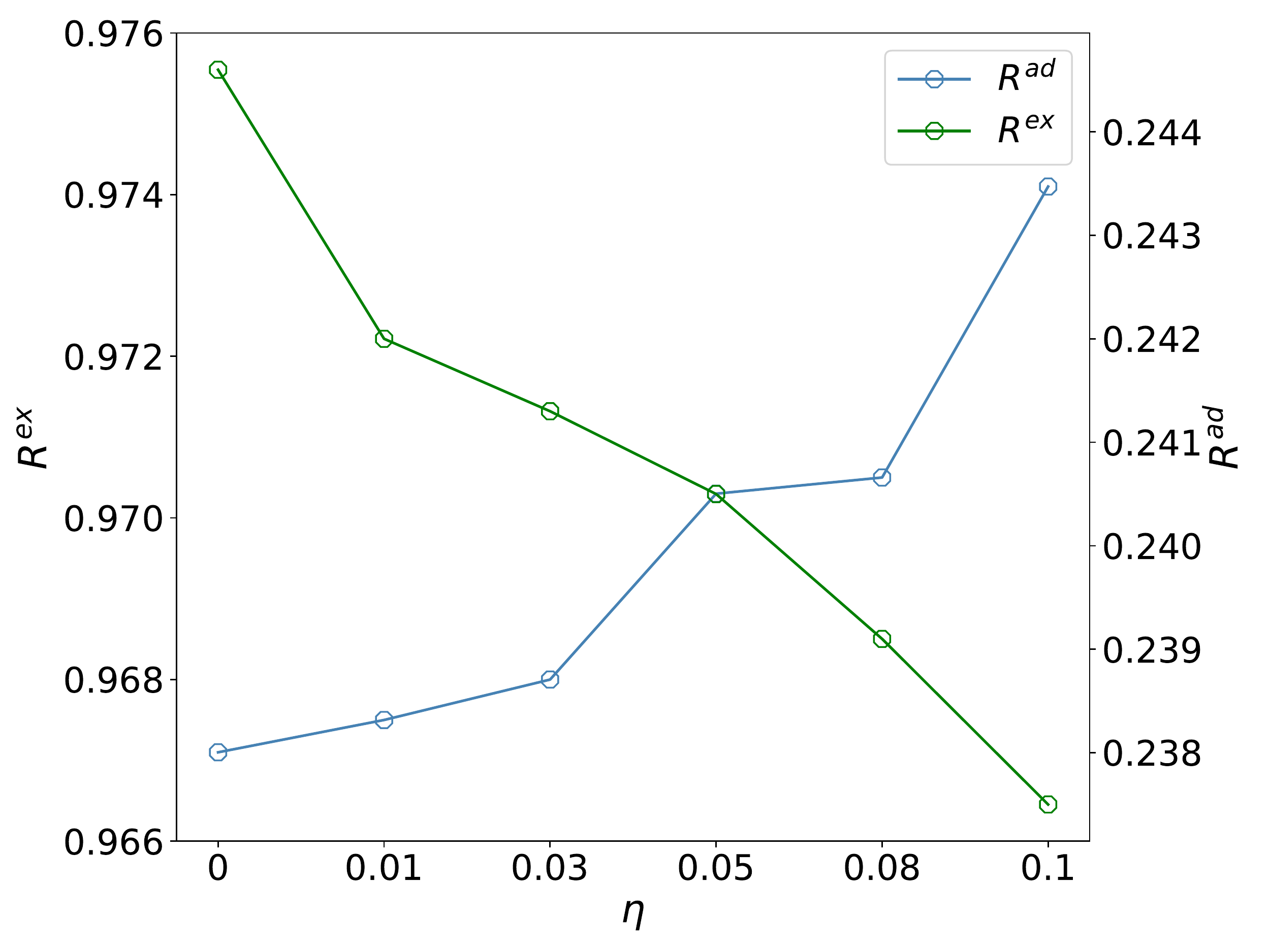} 
      \label{fig:subfig:ad} 
    \end{minipage}%
    } 
    \subfigure[{The Error bars of $\alpha$.}]{ 
      \centering
      \begin{minipage}{0.304\linewidth}
      \includegraphics[width=\textwidth]{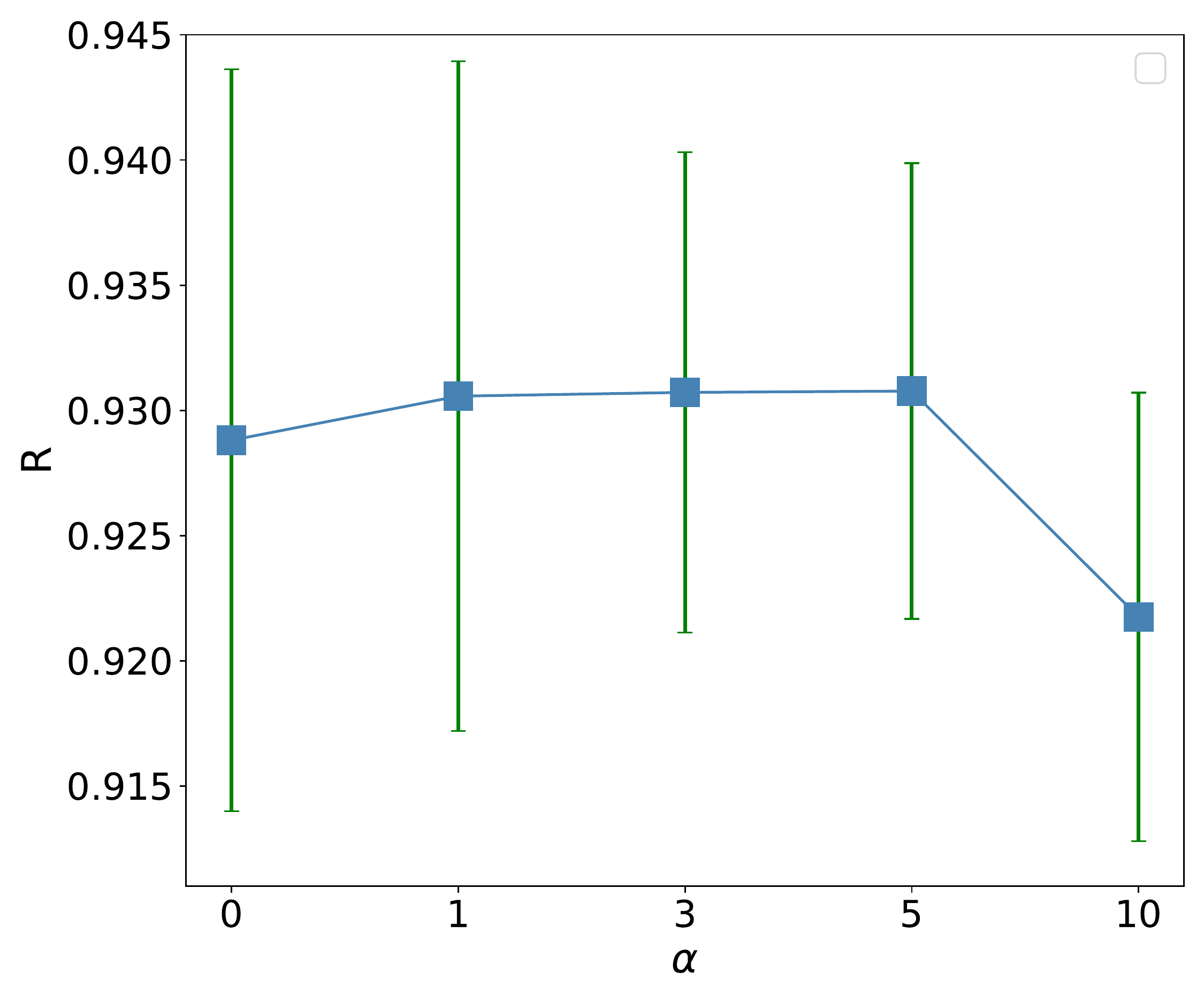} 
      \label{fig:subfig:fee} 
    \end{minipage}%
    } 
    \subfigure[{The Error bars of $\beta$.}]{ 
      \centering
      \begin{minipage}{0.303\linewidth}
      \includegraphics[width=\textwidth]{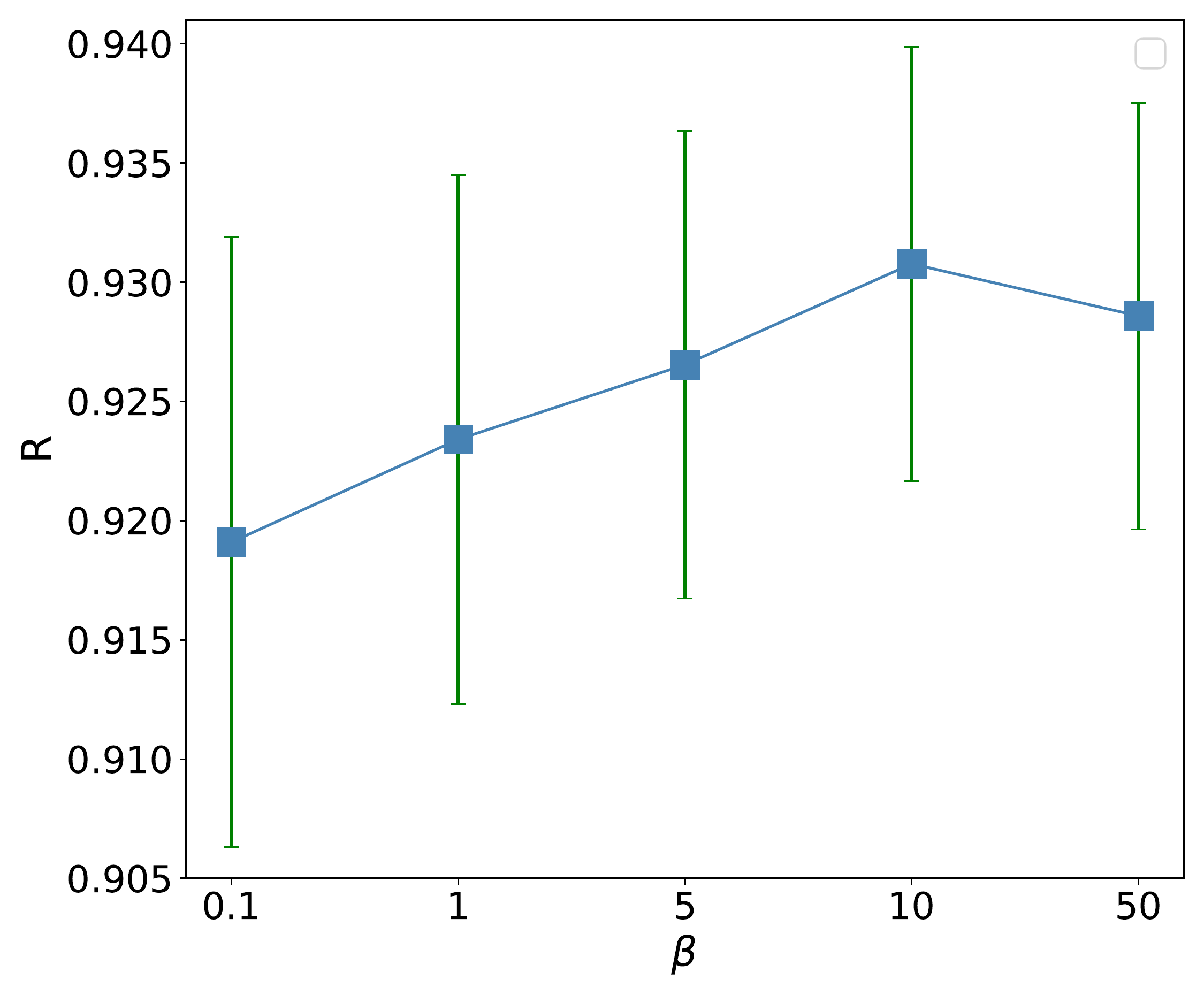}
      \label{fig:subfig:ctr} 
    \end{minipage}%
    } 
    \Description{The experimental results on the sensitivity of $\eta$, $\alpha$ and $\beta$.}
    \caption{{The experimental results on the sensitivity of $\eta$, $\alpha$ and $\beta$.}
    }
    \label{fig:eta}
  \end{figure*}

  \subsubsection{Evaluation Metrics}
  \red{
    We evaluate the model with revenue indicators and experience indicators.
  As for revenue indicators, we use ads revenue and service fee in a period to measure platform revenue. 
  Specifically, the ads revenue is gained from advertisers calculated using Generalized Second Price (GSP) \cite{edelman2007internet} and charged per click. The total ads revenue is calculated as $R^\text{ad}=\sum r^\text{ad}$.
  The service fee is charged from merchants' orders according to a certain percentage. and the total service fee is calculated as $R^\text{fee}=\sum r^\text{fee}$. 
  In our platform, user experience is measured by whether the user demand (e.g., finding a satisfying product) can be fulfilled. As for experience indicators, we use the average conversion rate and average experience score to measure user experience. The conversion rate calculated as $r^\text{cxr}=\sum{ \text{CTR} \times \text{CVR}}$ is the ratio of the number of orders placed by the user to the number of his/her requests. The experience score $r^\text{ex}$ defined in Section \ref{sec:problem} reflects the degree of satisfaction of the user demand.
  }
  
  \subsubsection{Hyperparameters}
  We implement Cross DQN with TensorFlow and apply a gird search for the hyperparameters. 
  The hidden layer sizes of the IRM are $(128, 64, 32, 8, 2)$ and the hidden layer sizes of the SDM are $(16, 8, 1)$. 
  The learning rate is $10^{-3}$, the optimizer is Adam \cite{kingma2014adam} and the batch size is $8,192$. 
  
\subsection{Offline Experiment}
In this section, we train Cross DQN with offline data and evaluate the performance using an offline estimator. 
Through extended engineering, the offline estimator models the user preference and aligns well with the online service.
We conduct experiments to answer the following two questions:
i) How does Cross DQN perform compared with other baselines? 
ii) How do different designs (e.g., SACU, \blue{MCAU}) and hyperparameter settings (e.g., $\alpha$, $\beta$) affect the performance of Cross DQN?
  
  \subsubsection{Baselines}
  We compare Cross DQN with the following five representative methods:
  \begin{itemize}
    \item \textbf{Fixed}. This method displays ads in fixed slots, such as the slot indexed by $3, 6, 9, \cdots$.
    \item \textbf{GEA} \cite{yan2020LinkedInGEA}. GEA is a non RL-based dynamic ads slots strategy. 
    It takes the impact of ads intervals into consideration and ranks the ads and organic items jointly with a rank score $\text{RS}=(\text{CTR} \times \emph{charge}+\text{GMV} \times \emph{takerate}) \exp(\beta' d)$, 
    {where \emph{charge} is the fee paid by advertisers, \emph{takerate} is the take rate (i.e., the fee charged on each transaction by the platform), 
    and $d$ is the interval between two ads.}
    \item \textbf{CTLRL} \cite{wang2019DDPGCHER}.
    Constrained Two-Level Reinforcement Learning (CTLRL)
    uses a two-level RL structure to allocate ads. 
    The upper level RL model decomposes the platform-level constraint into hour-level constraints, and the lower level RL model sets the hour-level constraint as the request-level constraint.
    \item \textbf{HRL-Rec} \cite{xie2021hierarchical}. 
    HRL-Rec is an RL-based dynamic ads slots strategy. It divides the integrated recommendation into two levels of tasks and solves using hierarchical reinforcement learning. Specifically, the model first decides the channel (i.e., select an organic item or an ad) and then determine the specific item for each slot.
    \item \textbf{DEAR} \cite{zhao2019deep}. 
    DEAR is also an RL-based dynamic ads slots strategy. 
    It designs a deep Q-network architecture to determine three related tasks jointly, i.e., i) whether to insert an ad to the recommendation list, and if yes, ii) the optimal ad and iii) the optimal location to insert.
  \end{itemize}

\subsubsection{Performance Comparison}
We present the experiment results under the same PAE level in Table \ref{tab:all_res} and have the following observations: 
i) Compared with all these baselines, Cross DQN achieves strongly competitive performance on both the revenue-related {metrics} and experience-related {metrics}. 
Specifically, Cross DQN improves over the best baseline w.r.t. $R^\text{ad}$, $R^\text{fee}$, $R^\text{cxr}$ and $R^\text{ex}$ by 3.09\%, 2.05\%, 0.83\% and 1.58\% separately. 
ii) Cross DQN outperforms the fixed slots strategy. 
A reasonable explanation is that the ads positions calculated by Cross DQN are more personalized, which leads to an increase in revenue as well as an improvement of user experience. 
iii) Cross DQN outperforms GEA, which indicates that an RL-based method may perform better than a rule-based method.
iv) Cross DQN also performs better than CTLRL possible due to the fact that 
the PAE of different requests of Cross DQN within the same hour are more personalized. 
v) Compared with the state-of-the-art RL-based methods, i.e., HRL-Rec and DEAR, the superior performance of Cross DQN justifies the explicit modeling of the arrangement signal.

 \subsubsection{Ablation Study}
 To verify the impact of different designs (SACU, \blue{MCAU}, batch-level constraint), we study three ablated variants of Cross DQN which have different components in SDM.
 \begin{itemize}
   \item Cross DQN (-aux) does not use the auxiliary loss of Cross DQN. Notice that without the help of the auxiliary loss, we can adjust the coefficients in the reward function to realize the same PAE as the PAE of other baselines.
   \item Cross DQN (-aux-\blue{mcau}) \blue{additionally blocks the MCAU and uses one self-attention unit instead} on top of the previous ablated version.
   \item Cross DQN (-aux-\blue{mcau}-sacu) concatenate the embeddings of the action and the state directly without SACU.
 \end{itemize}

The results shown in Table \ref{tab:all_res} reveal the following findings: 
i) The performance gap between Cross DQN (-aux-mcau-sacu) and Cross DQN (-aux-mcau) indicates the effectiveness of SACU.
By explicitly crossing the embeddings of the states and the actions, SACU can effectively \blue{generate cross matrix representation for subsequent extraction of arrangement signal}, therefore improving the overall metrics. 
ii) The MCAU is an additional process after the crossover to strengthen the mutual interaction. 
The performance of Cross DQN (-aux)
is superior to 
Cross DQN (-aux-mcau),
which verifies the effectiveness of \blue{extracting arrangement signal of different channel combinations}. 
iii) 
Cross DQN outperforms Cross DQN (-aux),
resulting from the fact that the batch-level constraint brings a certain revenue increase and makes the PAE in a period more stable.

\subsubsection{Hyperparameter Analysis}
We analyze 
the sensitivity of these three hyperparameters:
$\eta$, $\alpha$ and $\beta$. 
$\eta$ is the weight for the user experience in the reward function (cf. Eq. (\ref{eq:eta})). 
$\beta$ is the temperature parameter that controls the degree of the approximation in Eq. (\ref{eq:func_res}). 
$\alpha$ is the hyperparameter which balances the main loss and auxiliary loss (cf. Eq (\ref{eq:loss})).
 
\textbf{{Hyperparameter} $\eta$.} 
The experimental results of different values of $\eta$ are presented in Figure \ref{fig:eta}a. 
As $\eta$ increases, $R^\text{ex}$ increases but $R^\text{ad}$ decreases.
A reasonable explanation is that the system of dynamic ads allocation tends to insert fewer ads when $\eta$ becomes larger, which has a beneficial impact on user experience and fee.
 
\textbf{{Hyperparameter} $\alpha$.} As shown in Figure \ref{fig:eta}b, we find that the auxiliary loss for batch-level constraint has greater influence on return. When $\alpha$ increases, the standard deviation of reward decreases. This phenomenon shows that the PAE and revenue are more stable under batch-level constraint. It is worth noticing that the mean of reward increases when $\alpha$ changes from 0 to 1. One possible explanation is that the ads allocation under a certain batch-level constraint of PAE will be more reasonable, which ensures the quality of display results and improves the revenue and user experience. However, if $\alpha$ is too large, it will deviate from the learning goal, resulting in a decline in reward.
 
 \textbf{{Hyperparameter} $\beta$.} The right curve in Figure \ref{fig:eta} reveals that the mean of reward increases and the standard deviation of reward decreases as $\beta$ increases within a certain range. This phenomenon demonstrates that accurate calculation of PAE results in stable and high reward. On the contrary, the reward may decrease when $\beta$ exceeds a certain threshold, suggesting the necessity to carefully tune this parameter in practice.

 \subsection{Online Results}
 \label{result}
 We compare Cross DQN with fixed ads positions and both strategies are deployed on the Meituan platform through online A/B test. We keep total PAE the same for all methods for a fair comparison. As a result, we find that $R^\text{ad}$, $R^\text{fee}$ and $R^\text{ex}$ increase by 12.9\%, 10.2\% and 9.1\%, which demonstrates that our Cross DQN not only significantly increases the platform revenue, but also improves user experience. \blue{It is worth noting that this increase values are 11.5\%, 10.7\% and 10.0\% in offline experiments. One possible reason for this difference in absolute value is the differences in data distribution.}


 \section{Conclusion and Future Work}
 In this paper, we propose Cross DQN to optimize ads allocation in feed.
 In Cross DQN, we design State and Action Cross Unit and Multi-Channel Attention Unit to explicitly extract the arrangement signal that is the influence of the arrangement of items in mixed list on user behaviors.
 In addition, we introduce an auxiliary loss for batch-level constraint to achieve the personalization for different requests as well as the platform-level constraint. 
 Practically, both offline experiments and online A/B test have demonstrated the superior performance and efficiency of our solution.
 
In our scenario, user experience is also an important objective for the long-term growth of the platform since the improvement of user experience directly increases the retention rate and enhances the reputation of the platform.
In the future, it is beneficial to optimize for more user experience metrics and pay more attention to the modeling of long-term benefits.
In addition, it is worth noting that our method follows the offline reinforcement learning paradigm. 
Compared with online reinforcement learning, offline reinforcement learning faces additional challenges (such as the distribution shift problem). 
The impact of these challenges to the ads allocation problem is also a potential research direction in the future.

\bibliographystyle{ACM-Reference-Format}
\bibliography{main}

\appendix

\end{document}
\endinput